\documentclass{article}

\PassOptionsToPackage{numbers, compress}{natbib}

\usepackage[preprint]{neurips_2021}

\usepackage[utf8]{inputenc} 
\usepackage[T1]{fontenc}    
\usepackage{hyperref}       
\usepackage{url}            
\usepackage{booktabs}       
\usepackage{amsfonts}       
\usepackage{nicefrac}       
\usepackage{microtype}      
\usepackage{xcolor}         

\usepackage{times}
\usepackage{epsfig}
\usepackage{graphicx}
\usepackage{amsmath}
\usepackage{amssymb}
\usepackage{algorithm}
\usepackage{algorithmicx}
\usepackage{algpseudocode}
\usepackage{subfigure}
\usepackage{multirow}
\usepackage{diagbox}
\usepackage{amsthm}
\usepackage{bm}
\usepackage{caption}
\usepackage{enumitem}
\renewcommand{\mathbf}{\boldsymbol}

\def\x{\mathbf{x}}
\def\z{\mathbf{z}}

\usepackage{dsfont}

\newtheorem{assume}{Assumption}
\newtheorem{theorem}{Theorem}

\title{Semi-supervised Models are Strong Unsupervised Domain Adaptation Learners}
\author{%
  Yabin Zhang$^{1}$\\
   \And
   Haojian Zhang$^{2}$ \\
   \And
   Bin Deng$^{2}$ \\
   \And
   Shuai Li$^{1}$ \\
   \And
   Kui Jia$^{2}$ \\
   \And
   Lei Zhang$^{1}$ \\
   \AND
   $^1$ \textnormal{The Hong Kong Polytechnic University }
   \And
   $^2$ \textnormal{South China University of Technology}
}

\begin{document}
\newcolumntype{L}[1]{>{\raggedright\arraybackslash}p{#1}}
\newcolumntype{C}[1]{>{\centering\arraybackslash}p{#1}}
\newcolumntype{R}[1]{>{\raggedleft\arraybackslash}p{#1}}
\maketitle

\begin{abstract}		
	
		Unsupervised domain adaptation (UDA) and semi-supervised learning (SSL) are two typical strategies to reduce expensive manual annotations in machine learning. In order to learn effective models for a target task, UDA utilizes the available labeled source data, which may have different distributions from unlabeled samples in the target domain, while SSL employs few manually annotated target samples. Although UDA and SSL are seemingly very different strategies, we find that they are closely related in terms of task objectives and solutions, and SSL is a special case of UDA problems. Based on this finding, we further investigate whether SSL methods work on UDA tasks. By adapting eight representative SSL algorithms on UDA benchmarks, we show that SSL methods are strong UDA learners. Especially, state-of-the-art SSL methods significantly outperform existing UDA methods on the challenging UDA benchmark of DomainNet, and state-of-the-art UDA methods could be further enhanced with SSL techniques. We thus promote that SSL methods should be employed as baselines in future UDA studies and expect that the revealed relationship between UDA and SSL could shed light on future UDA development. Codes are available at \url{https://github.com/YBZh}. 
	

\end{abstract}

\section{Introduction}  \label{SecIntro}

Although deep models have achieved great successes on various tasks \cite{alexnet,rcnn}, they require a large amount of labeled training data and could not generalize to data of shifted distributions, impeding their applications in practical tasks. Given a new target task, the model learned from labeled source data of a previous task typically achieves degenerated performance on target data. It is time and labor consuming to collect a large amount of labeled training data for the new task. Fortunately, collecting unlabeled data is usually much easier; thus, there is a huge demand to learn efficient target models by utilizing labeled source data and unlabeled target data, falling in the realm of unsupervised domain adaptation (UDA) \cite{dann,transfer_survey}. 
Motivated by seminal UDA theories (e.g., Theorem \ref{The:ben-david}) \cite{ben2010theory, zhang2019bridging,zhang2020unsupervised}, which bound the target error with terms including domain discrepancy, several popular UDA methods \cite{dann,dan, mcd, cada,symnets} have been proposed to minimize the target risk by simultaneously learning domain-invariant representations and minimizing the source risk. 
Besides, the target risk is minimized by re-weighting labeled source samples in \cite{shimodaira2000improving, cortes2010learning}; pixel-level alignment is pursued in \cite{bousmalis2017unsupervised,liu2017unsupervised} with image style transfer techniques. A comprehensive survey on UDA methods can be found in \cite{wilson2020survey, da_survey} .


With similar motivations, semi-supervised learning (SSL) explores a different  way  from UDA to learn target models with minor manual efforts. Considering that manually annotating a few amount of target data is easy to achieve, SSL aims to learn target models from few labeled target data and a large amount of unlabeled target data \cite{van2020survey, ZhuSSLSurvey08, chapelle2009semi-old}.  
SSL methods are typically motivated by basic assumptions on the data structure, such as the smoothness assumption, low-density separation assumption, and manifold assumption \cite{van2020survey, ZhuSSLSurvey08, chapelle2009semi-old}. 
The smoothness assumption promotes that two samples close in the input space should present the same label, inspiring the consistency regularization \cite{temporal_ensembling, xie2019unsupervised} and graph-based methods \cite{zhou2004learning, jaakkola2002partially}. The low-density separation assumption states that the decision boundaries should lie in areas where few samples are observed, which motivates methods of entropy minimization \cite{sslem} and self-training \cite{lee2013pseudo}.  The manifold assumption suggests that the input space could be decomposed into multiple low-dimensional manifolds and samples on the same manifold should share the same label, whose representative method is the graph-based one \cite{zhou2004learning}.  Some methods simultaneously adopt multiple assumptions \cite{zhou2004learning,sohn2020fixmatch,mixmatch}. 

\begin{figure}
	\centering
	\subfigure[SSL] {
		\label{Fig:ssl_task}
		\centering
		\includegraphics[width=0.445\textwidth]{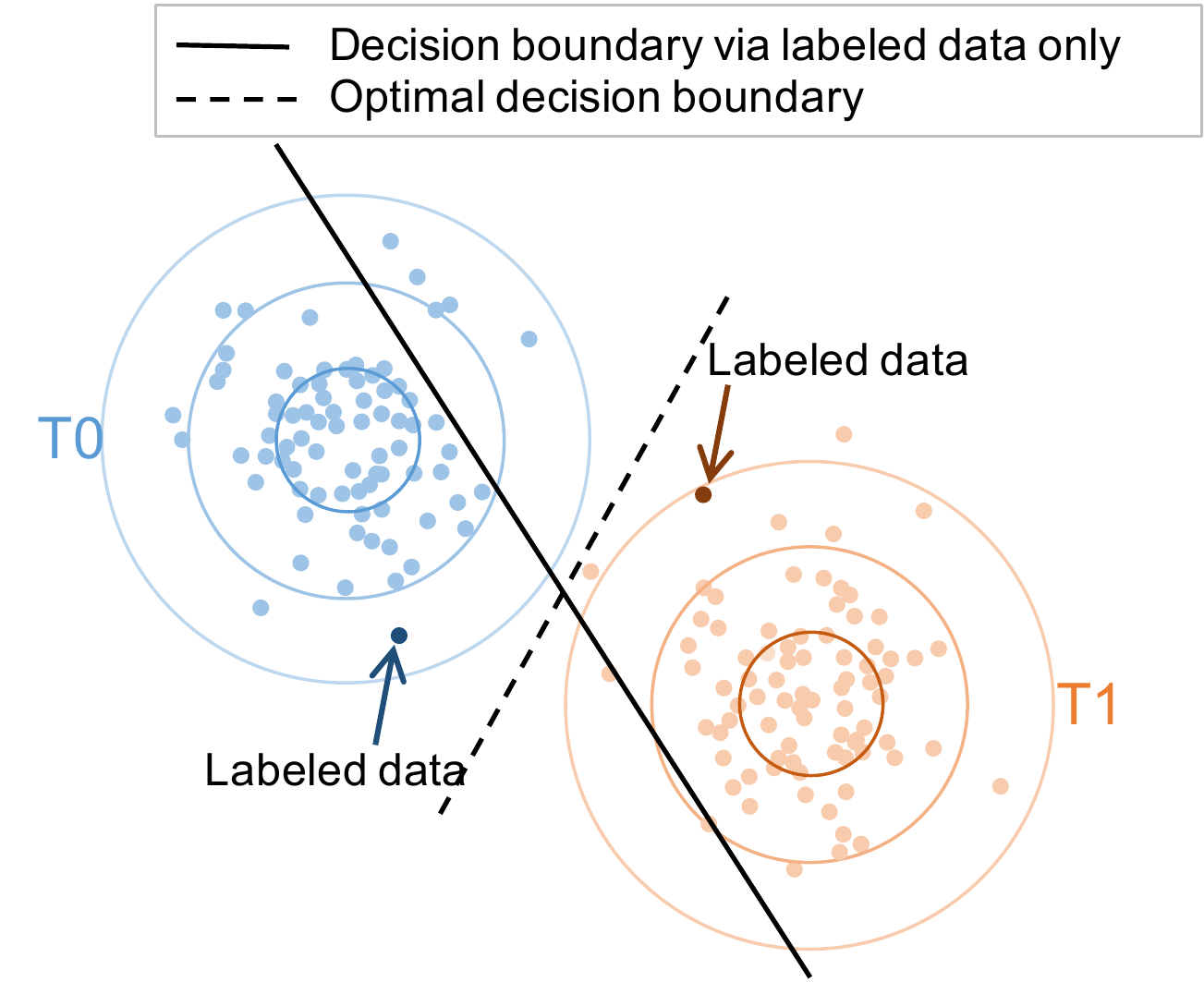}
	} \hfill
	\subfigure[UDA] {
		\label{Fig:uda_task}
		\includegraphics[width=0.515\textwidth]{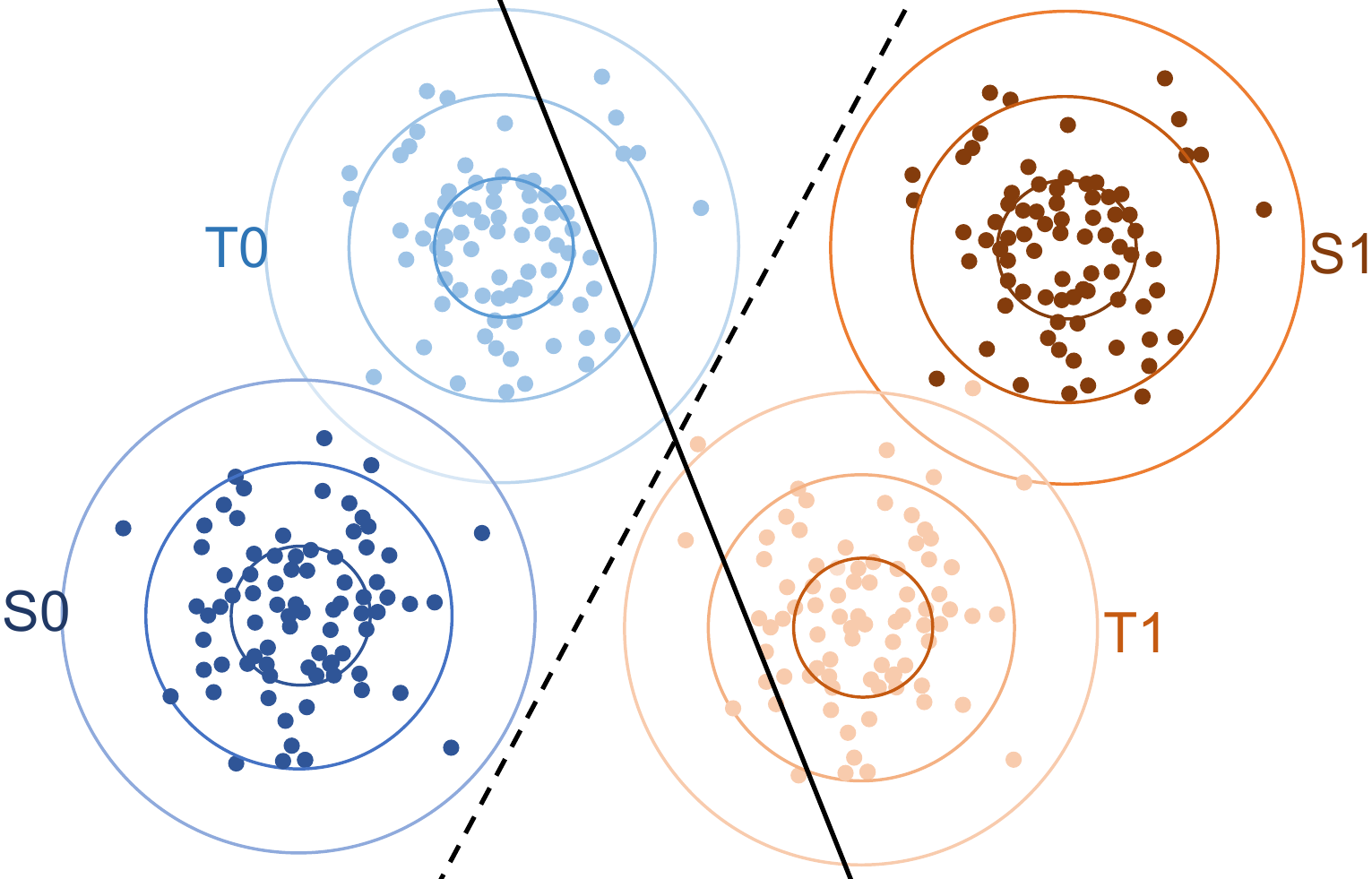}
	}
	\caption{Illustrations of (a) SSL and (b) UDA  on binary classification. In SSL, only a few samples per class are labeled and other training samples are unlabeled. In UDA, the source data (S$0$ and S$1$) are labeled and target data (T$0$ and T$1$) are unlabeled. Figure (a) is modified from Fig. 1 in \cite{van2020survey}. } \label{Fig:task_illustration}
\end{figure}

We illustrate the task settings of UDA and SSL in Figure \ref{Fig:task_illustration}. Although UDA and SSL are seemingly very different, we argue that they are closely related in the following aspects:
\begin{itemize}[leftmargin=0.4cm]
	\item UDA and SSL share a similar objective of utilizing unlabeled (target) data to improve the performance of the model trained with labeled (source) data only.
	\item Methods of UDA and SSL similarly utilize the underlying data structure.
	\item SSL is a special case of UDA problems when the target support covers source support. 
\end{itemize}
Based on the above relationships between UDA and SSL, we further investigate an important question: \textit{whether SSL methods work on UDA tasks? } To answer this question, we make in-depth analyses by applying eight representative SSL algorithms on four UDA benchmarks. Results clearly demonstrate that \textit{SSL methods are strong UDA learners}.
Especially, state-of-the-art SSL methods significantly outperform existing UDA methods on the challenging UDA benchmark of DomainNet, and state-of-the-art UDA methods could be further enhanced with SSL techniques.
Therefore, we argue that \textit{SSL methods, especially the state-of-the-art ones, should be employed as baselines in future UDA studies.}

\section{Background of UDA and SSL} \label{Sec:background}

In UDA, we are given $n_s$ labeled source samples $\mathcal{D}_s = \{ \x_s^i, y_s^i \}_{i=1}^{n_s}$ drawn from the  source  distribution $P_s(X,Y)$ and $n_t$ unlabeled target samples $\mathcal{D}_t = \{ \x_t^j \}_{j=1}^{n_t}$ drawn from the target distribution $P_t(X)$, where $X\in\mathcal{X}$ and $Y\in \mathcal{Y}=\{ 0, 1 \}$ (we consider the binary classification for simplicity). 
The unobserved labels in the target domain follow the distribution $P_t(Y|X)$, and the label space $\mathcal{Y}$ is shared across domains. 
The number of source samples $n_s$ and target samples $n_t$ are assumed to be large enough \cite{david2010impossibility}; thus the source data $\mathcal{D}_s$ and target data $\mathcal{D}_t$ could well represent  distributions of $P_s(X, Y)$ and  $P_t(X)$, respectively. The goal of UDA is to find a hypothesis $f: \mathcal{X} \to \{ 0,1\}$ to minimize the target risk $R_t(f) = \mathbb{E}_{(\x,y) \sim P_t(X,Y)} |f(\x) - y) |.$

By the convenience of deep models, we define the model as $f=h \circ g$, where $g: \mathcal{X} \to \mathcal{Z}$ lifts any input data in $\mathcal{X}$ to the feature space $\mathcal{Z}$ and $h: \mathcal{Z} \to \{ 0, 1 \}$ is the classifier defined in $\mathcal{Z}$. 
We denote the lifted distribution $P(g(X))$ as $P(Z)$.
We adopt the following covariate shift assumption\footnote{Although it is questionable whether $P_s(Y|g(X)) = P_t(Y|g(X))$ holds in the feature space \cite{gong2016domain, zhao2019learning}, $P_s(Y|X) = P_t(Y|X)$  in vanilla data space $\mathcal{X}$ generally holds in practical tasks.} for UDA problems:
\begin{assume} \label{Equ:covariate_shift}
	(Covariate shift) Domains $P_s(X,Y)$ and $P_t(X,Y)$ satisfy the covariate shift assumption if
	\begin{equation}
	P_s(X) \neq  P_t(X), \quad P_s(Y|X) = P_t(Y|X).
	\end{equation}
\end{assume}

 \textit{Importance-weighting}  \cite{shimodaira2000improving,cortes2010learning} and  \textit{learning domain-invariant feature representations} \cite{dann,dan} are two seminal UDA methods. Specifically, importance-weighting \cite{shimodaira2000improving} minimizes the target risk using importance-weighted source samples:
 \begin{equation} \label{Equ:method_iw}
 \mathcal{L}_{iw} = \frac{1}{n_s} \sum_{i=1}^{n_s} w(\x_s^i) \ell (f(\x_s^i), y_s^i),
 \end{equation}
 where $w(\x) = P_t(\x) / P_s(\x)$ and $\ell: \mathcal{R}^2 \to \mathcal{R}$ is the loss function. The efficacy of  importance-weighting in UDA problems is theoretically guaranteed  when the source support $supp(P_s)$ covers target support $supp(P_t)$ \cite{cortes2010learning}, where $supp(P) = \{ \x | P(\x) > 0\}$. 

Methods \cite{dann,dan, mcd,symnets}  aiming at learning domain-invariant feature representations are primarily motivated from the following UDA bound:
\begin{theorem} \label{The:ben-david}
	(Adaptation bound by Ben-David \emph{et al.} \cite{ben2010theory}) Given the hypothesis class $\mathcal{H}$, source distribution $P_s(Z,Y)$, and target distribution $P_t(Z,Y)$, the following inequality holds for all $h \in \mathcal{H}$,
	\begin{equation}
	R_t(h) \leq R_s(h) + \frac{1}{2} d_{\mathcal{H} \Delta \mathcal{H}}(P_s, P_t) + \lambda_{\mathcal{H}},
	\end{equation}
	where 
	\begin{align} 
	d_{\mathcal{H} \Delta \mathcal{H}}(P_s, P_t) = 
	2 \sup_{h,h'\in\mathcal{H}} \left| \mathbb{E}_{\z \in P_s(Z)} \mathbb{I}[h(\z) \neq h'(\z)]  - \mathbb{E}_{\z \in P_t(Z)} \mathbb{I}[h(\z) \neq h'(\z)]  \right|,
	\end{align}
	\begin{equation}
	\lambda_{\mathcal{H}} = \min_{h\in\mathcal{H}} \left[ R_t(h) + R_s(h)  \right].
	\end{equation}
	Here, the target risk is defined as $R_t(h) = \mathbb{E}_{(\z,y) \sim P_t(Z,Y)} |h(\z) - y |$ and the source risk  $R_s(h)$ is similarly defined. 
	 $\mathbb{I}[var] = \left\{ 
	\begin{array}{ll}
	1 & var = True\\
	0 & Others \\
	\end{array}\right.$. $\lambda_{\mathcal{H}}$ is a constant depending on the capacity of the hypothesis space $\mathcal{H}$.  
\end{theorem}
The bound in Theorem \ref{The:ben-david} is proposed for binary classification (i.e., $h(\z) \in \{ 0, 1\}$), which is extended to the multi-class setting in \cite{zhang2019bridging,zhang2020unsupervised}.

In SSL, labeled data $\mathcal{D}_l = \{\x_l^i, y_l^i \}_{i=1}^{n_l}$ and unlabeled data $\mathcal{D}_u = \{ \x_u^j \}_{j=1}^{n_u}$ are adopted as training data, where $\mathcal{D}_l$ and $\mathcal{D}_u$ are sampled from distributions of $P_{ssl}(X,Y)$ and $P_{ssl}(X)$, respectively.  It is typically assumed that $n_l \ll n_u$; in other words, the number of  unlabeled samples $n_u$ is assumed to be large enough to represent the distribution $P_{ssl}(X)$, while the number of labeled samples $n_l$ is too small to represent the distribution $P_{ssl}(X,Y)$.
The goal of SSL is to find a hypothesis $f: \mathcal{X} \to \{ 0,1\}$ to minimize the risk on the distribution $P_{ssl}(X,Y)$: $R_{ssl}(f) = \mathbb{E}_{(\x,y) \sim P_{ssl}(X,Y)} |f(\x) - y |.$  SSL methods are majorly motivated by assumptions on the data structure, as discussed in Section \ref{SecIntro}.

\section{Bridging UDA and SSL} \label{Sec:relationships}
\subsection{Bridging UDA and SSL: formulation and methodology } \label{Sec:similarity_methods}

We first introduce the general formulation of UDA and SSL, then we discuss their relationships in terms of methodology. Specifically, UDA and SSL both adopt the model trained with labeled (source) data as the baseline:
 \begin{equation}
 \min_{f=h \circ g} \mathcal{L}_{sup}(f, \mathcal{D}_l), 
 \end{equation} 
where $\mathcal{L}_{sup}(f, \mathcal{D}_l)$ is the supervised task loss and the labeled set $\mathcal{D}_l = \mathcal{D}_s$ in UDA. In both tasks, researchers introduce regularization terms with unlabeled data $\mathcal{D}_u$ (and labeled  data $\mathcal{D}_l$) as:
 \begin{equation} \label{Equ:regularization}
 \mathcal{L}_{reg}(f, \mathcal{D}_u, \mathcal{D}_l),
 \end{equation}
 where the unlabeled set $\mathcal{D}_u = \mathcal{D}_t$ in UDA.
In the regularization term (\ref{Equ:regularization}), only unlabeled (target) data are used  in \cite{jin2020minimum, sslem, lee2013pseudo} whereas other methods \cite{dann,sohn2020fixmatch, temporal_ensembling} additionally utilize the labeled (source) data.  Note that the regularization in (\ref{Equ:regularization}) is data-dependent, which is distinguished from the data-independent ones, e.g.,  weight decay \cite{krogh1992simple} and Dropout \cite{srivastava2014dropout}.
UDA and SSL both aim to utilize unlabeled (target) data to improve the performance of the model trained with labeled (source) data only, leading to the following objective\footnote{The transductive SSL methods, e.g., the graph-based ones \cite{zhou2004learning, belkin2004regularization}, also share a similar form of objective comprised of labeled data supervision and unlabeled data regularization; differently, these methods output label predictions of unlabeled data directly, instead of models. }:
\begin{equation} \label{Equ:general_objective}
\min_{f=h \circ g} \mathcal{L}_{sup}(f, \mathcal{D}_l) + \omega  \mathcal{L}_{reg}(f, \mathcal{D}_u, \mathcal{D}_l), 
\end{equation} 
where  $\omega$ is the trade-off parameter.

Although methods of UDA and SSL mentioned in Section \ref{SecIntro} are seemingly very different, they all follow the formulation (\ref{Equ:general_objective}). Further, many recent UDA methods adopt SSL techniques as the components, and some UDA methods are variants of SSL ones. 
For example, the entropy minimization \cite{sslem}, one typical SSL method, is used cooperatively with the domain discrepancy minimization in \cite{rtn,symnets,dirt}; label propagation \cite{zhou2004learning}, a graph-based SSL method, is also adopted in UDA methods \cite{ding2018graph, zhang2020label} together with the domain discrepancy minimization for performance boosting. Self-training, which is initially adopted in SSL \cite{lee2013pseudo}, and its variants are widely adopted in UDA methods \cite{zou2018unsupervised, zou2019confidence}.  Self-ensembling \cite{french2018selfensembling} is a variant of the SSL method of mean teacher \cite{tarvainen2017mean}, which boosts the vanilla mean teacher with strategies such as confidence threshold. 
Tang \emph{et al.} \cite{srdc} tackled UDA with the framework of cluster-then-label, which is also an SSL pipeline of a long history \cite{van2020survey,dara2002clustering}. 
MCC \cite{jin2020minimum} shares similar objectives to entropy minimization \cite{sslem} and self-training \cite{lee2013pseudo}, which promote the classification of low-density separation (i.e., low entropy predictions in \cite{sslem,lee2013pseudo} or less confusion predictions in \cite{jin2020minimum}). 

In the following section, we show that SSL is a special case of UDA problems. Although SSL techniques and their variants have been practically adopted in UDA methods, we explicitly reveal their relationships in Section \ref{Sec:special_case} and empirically illustrate the efficacy of vanilla SSL methods on UDA tasks in Section \ref{Sec:exp_ssl_baseline}; we promote that SSL methods should be employed as baselines in future UDA studies.

\subsection{SSL is a special case of UDA problems} \label{Sec:special_case}

As we discussed in Section \ref{Sec:background}, although labeled data $\{ \x_l^i, y_l^i\}_{i=1}^{n_l}$ are sampled from the distribution $P_{ssl}(X, Y)$ in SSL, they are typically insufficient to represent the overall $P_{ssl}(X, Y)$ since very few labeled data are sampled (i.e., $n_l$ is small); in other words, the few labeled data $\{ \x_l^i, y_l^i \}_{i=1}^{n_l}$ could only represent part of the overall distribution $P_{ssl}(X, Y)$, i.e., a sub-domain of $P_{ssl}(X, Y)$. Among all possible sub-domains, the one owning the smallest support set could be represented as follows:
\begin{align}
P_{small}(X):  P_{small}(X=\x) >0 \iff \x \in \{ \x_l^i \}_{i=1}^{n_l},   \quad 
P_{small}(Y|X) = P_{ssl}(Y|X)
\end{align}

With $P_{small}$ and $P_{ssl}$ as distributions of labeled and unlabeled data respectively and the covariate shift assumption  \ref{Equ:covariate_shift} in UDA, the SSL task turns to a UDA problem, where the unlabeled target support $supp(P_{ssl})$ covers the labeled source support $supp(P_{small})$, i.e., $supp(P_{small}) \subset supp(P_{ssl})$. 

\begin{figure}
	\centering
	\includegraphics[width=0.9\textwidth]{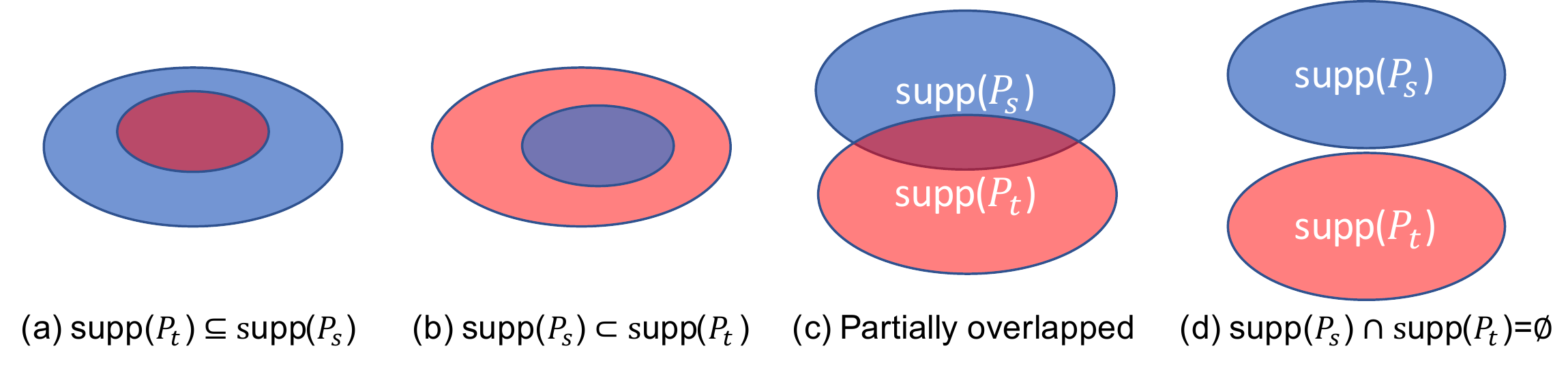}
	\caption{An illustration of the four UDA settings with different relationships between the source support $supp(P_s)$ and target support $supp(P_t)$. The settings of (a) $supp(P_t) \subseteq supp(P_s)$ and (b) $supp(P_s) \subset supp(P_t)$ are respectively investigated with the importance weighting \cite{shimodaira2000improving, cortes2010learning} and SSL; the other two settings (c) and (d) are not specifically studied yet to the best of our knowledge.  } \label{Fig:four_setting}
\end{figure}

\noindent \textbf{Remark 1.} UDA problems could be classified into four settings according to the relationship between the source support $supp(P_s)$ and target support $supp(P_t)$, as illustrated in Figure \ref{Fig:four_setting}. Specifically, the importance weighting algorithm (\ref{Equ:method_iw})  \cite{shimodaira2000improving, cortes2010learning} is proposed to tackle the specific UDA setting where $supp(P_t) \subseteq supp(P_s)$. When $supp(P_s) \subset supp(P_t)$, this UDA problem could be solved by SSL methods since SSL tasks coincidentally fall into the realm of this UDA setting.  Another two settings, i.e., $supp(P_s) \cap supp(P_t) = \varnothing$ and the remaining partially overlapped setting, are not specifically investigated yet to the best of our knowledge. 
Although the SSL task falls in a specific UDA setting of $supp(P_s) \subset supp(P_t)$, the general assumptions of smoothness, low-density separation, and manifold in SSL methods (cf. Section \ref{SecIntro}) commonly hold in general UDA tasks, supporting the empirical successes of SSL methods on practical UDA tasks, as will be demonstrated  in Section \ref{Sec:exp_ssl_baseline}.


\noindent \textbf{Remark 2.} We discuss the potential impact of our research on the UDA development. Motivated by seminal UDA theories \cite{ben2010theory,zhang2019bridging,zhang2020unsupervised}, the most popular UDA framework minimizes the target risk by simultaneously learning domain-invariant representations and minimizing the source risk \cite{dann,cada,mcd,symnets}.  Although some successes have been achieved, this framework has some limitations. 
As empirically studied in \cite{chen2019transferability}, the discrimination of target data may be deteriorated when learning domain-invariant feature representations via adversarial training \cite{dann}. Zhao \emph{et al.} \cite{zhao2019learning} theoretically indicated that learning domain-invariant representations and minimizing the source risk are not sufficient to guarantee a small target risk, since the minor joint error $\lambda_{\mathcal{H}}$ in Theorem (\ref{The:ben-david}) may be enlarged as the change of the feature space $\mathcal{Z}$. 
Zhao \emph{et al.} \cite{zhao2019learning}  emphasized that the conditional shift across domains plays an important role in the target risk minimization under the UDA framework \cite{ben2010theory, zhang2019bridging,zhang2020unsupervised}, but how to practically minimize the conditional shift is unclear due to the missing of target labels. 
To minimize the conditional shift, one may turn to pseudo labels of target data \cite{xie2018learning,can}, but these pseudo labels may be unreliable. 
Different from the seminal UDA framework, where unlabeled target data are utilized to explicitly minimize the domain divergence, recent UDA methods have been proposed to explore and utilize the data structure of unlabeled data \cite{jin2020minimum,french2018selfensembling,srdc}, which is closely related to SSL assumptions on the data structure. We expect that the revealed relationship between UDA and SSL could promote the development of UDA research in terms of both algorithms and theories.

\textbf{A toy example.} 
To facilitate the understanding of our findings, we construct a simple $1$-dimensional example (i.e., $\mathcal{X} = \mathbb{R}$) to intuitively illustrate that the SSL task could be formulated as a UDA problem. 
With $U(a,b)$ as the uniform distribution within range $[a,b]$, where $a < b$, we introduce the target distribution as:
\begin{equation}
P_{ssl}(X) = U(0,1), \quad P_{ssl}(Y=1|X=\x) = \left\{ 
\begin{array}{rl}
0 & \x \leq 0.5 \\
1 & \x > 0.5 
\end{array} \right..
\end{equation}
We suppose only one labeled sample per-category is given in SSL\footnote{The same conclusion holds for multiple, but few,  samples per-category in multiclass classification; we analyze the binary case with one labeled sample per-category for simplicity.}. Specifically, the two labeled samples are $\mathcal{D}_l = \left\{  (\x_l^1 = c, y_l^1 =0 ), (\x_l^2 = d, y_l^2 =1) \right\}$, where $0 \leq c \leq 0.5$ and $0.5 < d \leq 1$.
Then we could introduce $P_{small}$, the possible distribution for $\mathcal{D}_l$ with the smallest support set, as a variant of the Bernoulli distribution:
\begin{equation}
P_{small}(X): P_{small}(X=\x)= \left\{ 
\begin{array}{ll}
0.5 & \x = c \\
0.5 & \x = d \\
0 & Others
\end{array}\right., P_{small}(Y|X) = P_{ssl}(Y|X).
\end{equation}
With $P_{small}$ and $P_{ssl}$ as distributions of source and target domains respectively, the SSL task is formulated as a UDA problem, as illustrated in Figure \ref{Fig:ss_da_unify}.

\begin{minipage}{0.48\textwidth}
	\includegraphics[width=0.95\textwidth]{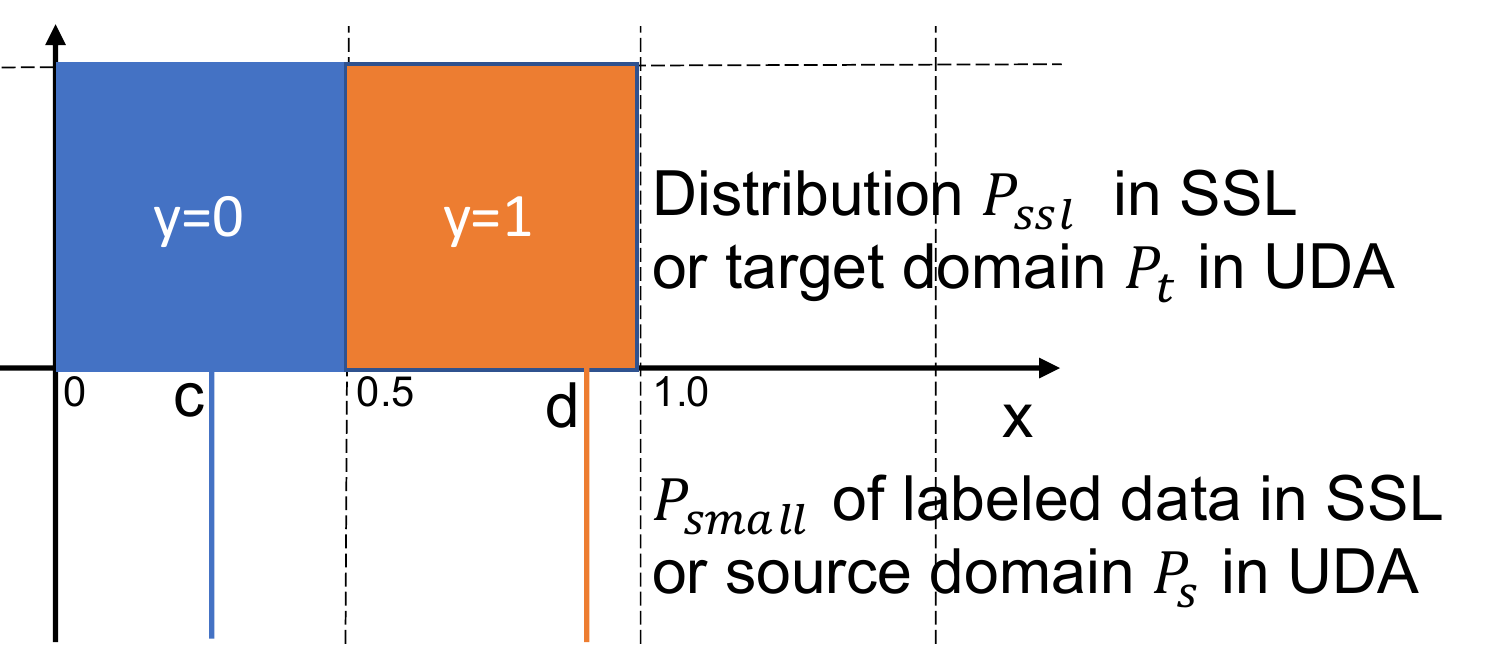} 
	\captionof{figure}{A toy example bridging SSL and UDA.} \label{Fig:ss_da_unify}
\end{minipage}
\begin{minipage}{0.45\textwidth}
	\centering 
	\begin{tabular}{l|ccc}
		\hline
		\multirow{2}{*}{Datasets} & Num. of & Num. of  & Num. of \\
		&   Domains &  Classes & Samples \\
		\hline
		Office31 \cite{saenko2010adapting} & 3 & 31 & 4.1K \\
		OfficeHome \cite{venkateswara2017deep} & 4 & 65 & 15.5K \\
		VisDA-2017 \cite{peng2017visda} & 2 & 12 & 280.2K \\
		DomainNet \cite{peng2019moment} & 6 & 345 & 586.6K \\
		\hline
	\end{tabular}
	\captionof{table}{Summary of adopted datasets. } \label{Tab:datasets}
\end{minipage}

\begin{table}
	\centering
	\caption{Transductive results on the small-scale Office31 dataset (ResNet50).} \label{Tab:office31}
	\begin{small}
		\begin{tabular}{l|l|cccccc|c}
			\hline
			&Methods & A$\to$W & D$\to$W & W$\to$D & A$\to$D & D$\to$A & W$\to$A & Avg. \\
			\hline
			&Source Only &  77.3$\pm$1.1 & 95.5$\pm$0.5 & 99.1$\pm$0.1 & 81.4$\pm$0.4 & 64.1$\pm$0.1 & 65.3$\pm$0.7 & 80.5  \\
			\hline
			\multirow{7}{*}{\centering\rotatebox{90}{UDA}} 
			&DANN \cite{dann} & 83.9$\pm$2.1 & 97.3$\pm$0.2 & \textbf{100.0}$\pm$.0 & 81.9$\pm$0.8 & 72.5$\pm$0.1 & 72.8$\pm$0.1 & 84.7 \\
			&MCD \cite{mcd} & 88.1$\pm$1.5 & 98.3$\pm$0.2 & \textbf{100.0}$\pm$.0 & 87.5$\pm$0.5 & 71.6$\pm$0.8 & 68.1$\pm$0.1 & 85.6 \\
			&CDAN \cite{cada} & 89.3$\pm$2.6 & 98.6$\pm$0.5 & \textbf{100.0}$\pm$.0 & 94.1$\pm$1.8 & 75.5$\pm$2.7 & 71.9$\pm$0.8 & 88.2 \\
			& AFN \cite{xu2019larger} &  92.3$\pm$0.7 & \textbf{99.0}$\pm$0.1 & \textbf{100.0}$\pm$.0 & 94.9$\pm$1.6 & 73.6$\pm$0.2 & 70.7$\pm$0.8 & 88.3 \\
			&MDD \cite{zhang2019bridging} & 89.2$\pm$1.3 & 98.7$\pm$0.4 & \textbf{100.0}$\pm$.0 & 94.4$\pm$2.3 & 76.0$\pm$1.0 & 73.8$\pm$0.1 & 88.7 \\
			\cline{2-9}
			& Self-ensembling \cite{french2018selfensembling} &  86.1$\pm$0.5 & 98.3$\pm$0.2 & \textbf{100.0}$\pm$.0 & 89.2 $\pm$0.3 & 68.8$\pm$0.1 & 67.5$\pm$0.5 & 85.0 \\
			& MCC \cite{jin2020minimum} & 93.0$\pm$0.1 & 98.1$\pm$0.1 & 99.8$\pm$0.2 & 95.0$\pm$0.8 & 76.6$\pm$0.1 & 76.1$\pm$0.6 & 89.8 \\
			\hline
			\multirow{8}{*}{\centering\rotatebox{90}{SSL}} 
			& VAT \cite{vat} & 81.8$\pm$0.7 & 98.5$\pm$0.2& 99.9$\pm$0.1 & 82.5$\pm$1.0   &  66.2$\pm$0.6 & 65.3$\pm$1.0 & 82.4 \\
			& Mean Teacher \cite{tarvainen2017mean} &83.7$\pm$0.2 & 98.6$\pm$0.2 &  \textbf{100.0}$\pm$.0 & 84.1$\pm$0.2 & 68.6$\pm$0.3&  66.7$\pm$0.2 & 83.6 \\
			& $\pi$-Model \cite{temporal_ensembling} & 87.7$\pm$0.6 & 98.9$\pm$0.1 &  \textbf{100.0}$\pm$.0 & 87.1$\pm$0.7 & 69.6$\pm$0.5 & 68.0$\pm$0.1 &  85.2 \\
			& MixMatch \cite{mixmatch} & 89.3$\pm$0.5 & \textbf{99.0}$\pm$0.1 & 99.8$\pm$0.2 & 90.5$\pm$0.3 & 68.6$\pm$0.2 &  67.8$\pm$0.3 & 85.8 \\
			& Self-training \cite{lee2013pseudo} & 88.0$\pm$0.2 & 98.8$\pm$0.2 & \textbf{100.0}$\pm$.0 & 90.6$\pm$0.2 & 69.5$\pm$0.1 & 68.6$\pm$0.2 & 85.9 \\
			& Entropy mini. \cite{sslem} & 89.3$\pm$0.6 & 98.9$\pm$0.1  &  \textbf{100.0}$\pm$.0 & 88.0$\pm$0.2 & 72.7$\pm$0.6  & 68.2$\pm$0.4 & 86.2 \\
			& FixMatch \cite{sohn2020fixmatch} & 92.6$\pm$0.3 & 98.9$\pm$0.3 & \textbf{100.0}$\pm$.0 & 93.3$\pm$0.5 & 73.5$\pm$1.0 & 71.6$\pm$0.1 & 88.3 \\
			& Xie \emph{et al.} \cite{xie2019unsupervised} & 93.6$\pm$0.8 & 98.6$\pm$0.4 & \textbf{100.0}$\pm$.0 & \textbf{95.6}$\pm$0.2 & 73.5$\pm$0.3 & 74.2$\pm$0.2 & 89.2   \\
			\hline
			& MDD + Consistency & 92.2$\pm$0.4 & 97.9$\pm$0.1 & \textbf{100.0}$\pm$.0 & 93.8$\pm$0.6 & 75.0$\pm$0.1 & 75.0$\pm$0.2 & 89.0 \\
			& MCC + Consistency & \textbf{94.0}$\pm$0.7 & 98.1$\pm$0.1 & 99.6$\pm$0.2 & 93.6$\pm$0.4 & \textbf{77.3}$\pm$0.2 & \textbf{76.7}$\pm$0.1 & \textbf{89.9}  \\
			\hline
		\end{tabular}
	\end{small}
\end{table}

\begin{table}
	\centering
	\caption{Transductive results on the medium-scale OfficeHome dataset (ResNet50). } \label{Tab:officehome}
	\begin{small}
		\begin{tabular}{l|l|cccccc|c}
			\hline
			&Methods & Ar$\to$Cl & Ar$\to$Rw & Cl$\to$Pr & Pr$\to$Ar & Pr$\to$Rw & Rw$\to$Cl & Avg. \\
			\hline
			 &Source Only & 42.6$\pm$0.3 & 74.1$\pm$0.4 & 61.7$\pm$0.4 & 54.5$\pm$0.1 & 72.8$\pm$0.1 & 44.2$\pm$0.1 & 58.3 \\
			\hline
			\multirow{7}{*}{\centering\rotatebox{90}{UDA}} 
			& AFN \cite{xu2019larger} & 48.3$\pm$0.1 & 76.1$\pm$0.1 & 68.5$\pm$0.6 & 62.6$\pm$0.1 & 76.1$\pm$0.1 & 52.6$\pm$0.1 & 64.0 \\ 
			&DANN \cite{dann} & 50.5$\pm$0.7 & 75.1$\pm$0.1 & 65.0$\pm$0.1 & 61.2$\pm$0.4 & 78.5$\pm$0.1 & 56.5$\pm$0.1 & 64.5 \\
			&MCD \cite{mcd} & 50.8$\pm$0.8 & 77.4$\pm$0.1 & 69.3$\pm$0.1 & 61.7$\pm$0.4 & 78.2$\pm$0.1 & 56.9$\pm$0.4 & 65.7 \\
			&CDAN \cite{cada}  & 51.2$\pm$0.1 & 79.3$\pm$0.1 & 70.3$\pm$0.8 & 65.4$\pm$0.1 & 80.8$\pm$0.5 & 57.3$\pm$0.4 & 67.4 \\
			&MDD \cite{zhang2019bridging} & 54.2$\pm$0.4 & 80.0$\pm$0.2 & 72.3$\pm$0.1 & 66.1$\pm$0.1 & 80.0$\pm$0.3 & 58.5$\pm$0.5 & 68.5 \\
			\cline{2-9}
			& Self-ensembling \cite{french2018selfensembling} & 47.2$\pm$0.1 & 75.9$\pm$0.4 & 67.8$\pm$0.5 & 62.4$\pm$2.1 & 76.2$\pm$0.6 & 53.1$\pm$1.1 & 63.8 \\
			& MCC \cite{jin2020minimum} & 55.5$\pm$0.1 & 83.0$\pm$0.6 & 75.8$\pm$0.5 &  \textbf{69.2}$\pm$0.2 & 81.9$\pm$0.1 & 59.4$\pm$0.8 & 70.8 \\
			\hline
			\multirow{8}{*}{\centering\rotatebox{90}{SSL}} 
			& Mean Teacher \cite{tarvainen2017mean} & 44.3$\pm$0.2 & 75.0$\pm$0.1 &  62.1$\pm$0.1 & 54.7$\pm$0.4 & 72.9$\pm$0.2 & 45.0$\pm$0.2 & 59.0 \\
			& VAT \cite{vat} & 44.0$\pm$0.1 & 74.8$\pm$0.2 & 62.3$\pm$0.1 & 55.0$\pm$0.3 & 72.8$\pm$0.1 & 45.1$\pm$0.2 & 59.0 \\
			& $\pi$-Model \cite{temporal_ensembling} & 44.1$\pm$0.5 & 76.0$\pm$0.1 & 64.1$\pm$0.1 & 54.2$\pm$0.3 & 73.7$\pm$0.1 & 45.0$\pm$0.3 & 59.4 \\
			& Self-training \cite{lee2013pseudo} & 46.2$\pm$0.3 & 76.4$\pm$0.1 & 66.1$\pm$0.1 & 58.0$\pm$0.2 & 74.9$\pm$0.1 & 48.8$\pm$0.1 & 61.7 \\
			& MixMatch \cite{mixmatch} & 45.3$\pm$0.2 & 76.6$\pm$0.1 & 71.1$\pm$0.1 & 60.1$\pm$0.1 & 77.0$\pm$0.2 & 50.8$\pm$0.2 & 63.5  \\
			& Entropy mini. \cite{sslem} & 52.0$\pm$0.2 & 77.1$\pm$0.2 & 71.3$\pm$0.1 & 61.4$\pm$0.1 & 78.5$\pm$0.1 & 54.3$\pm$0.1 &  65.8\\
			& Xie \emph{et al.} \cite{xie2019unsupervised} & 52.1$\pm$1.8 & 79.4$\pm$0.1 & 72.3$\pm$0.3 & 65.2$\pm$0.3 & 79.2$\pm$0.2 & 56.9$\pm$0.5 &  67.5  \\
			& FixMatch \cite{sohn2020fixmatch} & 52.5$\pm$0.8 & 79.4$\pm$0.2 & 73.6$\pm$0.1 & 66.5$\pm$0.6 & 80.0$\pm$0.1 & 57.3$\pm$0.5 & 68.2 \\
			\hline
			& MDD + Consistency & \textbf{58.2}$\pm$0.2 & 81.2$\pm$0.2 & 77.0$\pm$1.4 & 69.1$\pm$0.3 & 82.1$\pm$0.2 & 60.0$\pm$0.3 & 71.2  \\
			& MCC + Consistency & 57.6$\pm$0.2 &  \textbf{84.1}$\pm$0.3 &  \textbf{78.3}$\pm$0.1 &  \textbf{69.2}$\pm$0.2 &  \textbf{83.3}$\pm$0.3 &  \textbf{60.8}$\pm$0.1 & \textbf{72.2}  \\
			\hline
		\end{tabular}
	\end{small}
\end{table}

\begin{table}
	\centering
	\caption{Transductive and inductive results on the large-scale VisDA-2017 dataset (ResNet101). } \label{Tab:visda}
	\begin{tabular}{l|l|cc||cc}
		\hline
		&\multirow{2}{*}{Methods} & \multicolumn{2}{c||}{Transductive} & \multicolumn{2}{c}{Inductive}\\
		& & Acc. over Ins.  & Acc. over Cate.  & Acc. over Ins.  & Acc. over Cate. \\
		\hline
		&    Source Only & 55.4$\pm$1.1 & 51.0$\pm$0.8 & 52.3$\pm$2.1 & 50.4$\pm$1.6 \\
		\hline
		\multirow{6}{*}{\centering\rotatebox{90}{UDA}} 
		&DANN \cite{dann} &74.1$\pm$0.7 & 79.1$\pm$0.6 & 72.0$\pm$1.2 & 71.3$\pm$1.1 \\
		&MCD \cite{mcd} & 76.8$\pm$0.6 & 78.1$\pm$0.9 & 73.3$\pm$0.1 & 71.5$\pm$0.0 \\
		&CDAN \cite{cada} & 77.3$\pm$0.6 & 81.3$\pm$0.6 & 74.6$\pm$1.0 & 73.5$\pm$1.3 \\
		& AFN \cite{xu2019larger} & 75.5$\pm$0.1 & 76.0$\pm$0.3 & 74.4$\pm$0.6 &  73.9$\pm$2.7 \\ 
		&MDD \cite{zhang2019bridging} & 77.8$\pm$0.8 & 80.9$\pm$0.4 & 75.0$\pm$0.2 & 73.7$\pm$0.2   \\
		\cline{2-6}
		& Self-ensembling \cite{french2018selfensembling} & 80.2$\pm$0.9 & 80.3$\pm$1.3 & 74.4$\pm$0.6 & 75.0$\pm$0.6 \\
		& MCC \cite{jin2020minimum} & 80.3$\pm$0.4 & 83.3$\pm$0.4 & 78.5$\pm$0.5 & 78.3$\pm$0.4 \\ 
		\hline
		\multirow{8}{*}{\centering\rotatebox{90}{SSL}}  
		& $\pi$-Model \cite{temporal_ensembling} & 65.4$\pm$0.3 & 60.4$\pm$0.2 &  59.6$\pm$0.1 & 57.5$\pm$0.4 \\
		& VAT \cite{vat}& 62.4$\pm$0.2 & 62.3$\pm$0.2 & 62.5$\pm$0.5 & 60.1$\pm$0.4 \\
		& Mean Teacher \cite{tarvainen2017mean} & 63.0$\pm$0.5 & 59.8$\pm$0.7 & 63.4$\pm$0.6 & 60.7$\pm$0.6 \\
		& Entropy mini. \cite{sslem}& 74.9$\pm$0.2 & 74.8$\pm$0.1  & 69.4$\pm$0.3  & 66.1$\pm$0.3 \\
		& Self-training \cite{lee2013pseudo} & 75.2$\pm$0.2 & 74.0$\pm$0.3 & 73.1$\pm$0.1 &  70.5$\pm$0.1 \\
		& MixMatch \cite{mixmatch} & 78.4$\pm$0.6 & 78.9$\pm$0.7 & 74.8$\pm$0.6 & 73.6$\pm$0.8  \\
		& Xie \emph{et al.} \cite{xie2019unsupervised} & 79.0$\pm$0.7 & 79.0$\pm$0.9 & 76.7$\pm$0.7 & 75.7$\pm$1.3 \\
		& FixMatch \cite{sohn2020fixmatch}  & 80.5$\pm$0.1 & 81.1$\pm$0.2 & 78.6$\pm$0.5 & 77.9$\pm$0.8 \\
		\hline
		& MDD + Consistency & 78.9$\pm$1.0 & 76.9$\pm$3.0 & 75.4$\pm$3.5 & 76.4$\pm$3.9   \\
		& MCC + Consistency & \textbf{82.6}$\pm$0.4 & \textbf{85.0}$\pm$0.5 & \textbf{82.8}$\pm$0.8 & \textbf{83.1}$\pm$0.8   \\
		\hline
		& Oracle & -- & --  & 88.2$\pm$0.1 &  88.2$\pm$0.2   \\
		\hline
	\end{tabular}
\end{table}

\section{Experiments} \label{SecExp}
The adopted datasets are briefly summarized in Table \ref{Tab:datasets} and detailed in the appendices. 
For datasets of VisDA-2017 and DomainNet, we adopt the source and target training sets for training and report the performance on the target training set and target test set in the transductive UDA setting and inductive UDA setting, respectively. 
For datasets of Office31 and OfficeHome, we adopt all data in source and target domains for training and report the results on target data in the transductive UDA setting following the standard setup \cite{dann}. 
We report the classification accuracy over instances on all datasets and additionally report the accuracy over categories on the VisDA-2017 dataset.  

We adopt the public library \cite{dalib}, which is licensed under the MIT License, to implement the Source Only, DANN \cite{dann}, Self-ensembling \cite{french2018selfensembling}, CDAN \cite{cada}, MCD \cite{mcd}, MDD \cite{zhang2019bridging}, AFN \cite{xu2019larger}, and MCC \cite{jin2020minimum}, since models in \cite{dalib} are well-tuned and achieve superior results over official implementations.  In Oracle settings of VisDA-2017 and DomainNet, we fine-tune the model with labeled target training data and report results on target test data, which sets the upper bound.  We report results of $mean\pm std$ with  experiments of three random seeds.  We run all experiments with eight RTX-2080Ti GPUs.

\subsection{Strong UDA baselines with SSL methods} \label{Sec:exp_ssl_baseline}

To employ SSL methods on UDA tasks, we set $\mathcal{D}_l = \mathcal{D}_s$ and $\mathcal{D}_u = \mathcal{D}_t$. 
The adopted SSL methods \cite{lee2013pseudo,sslem,vat,temporal_ensembling,tarvainen2017mean,mixmatch,xie2019unsupervised,sohn2020fixmatch} are briefly introduced in the appendices. 
We adopt the popular network architecture and optimization strategies \cite{dann} in all SSL methods for all UDA tasks:
we adopt a ResNet \cite{resnet} pre-trained on the ImageNet dataset \cite{deng2009imagenet} as the feature extractor $g$ after removing the last fully connected (FC) layer, and add a new task classifier $h$ of one FC layer; 
the overall model is updated by stochastic gradient descent (SGD) with a momentum of $0.9$; the learning rate is adjusted by $\eta_p = \frac{0.01}{(1+10 p)^{0.75}}$, where $p$ is the progress of training iterations linearly changing from $0$ to $1$;
the learning rate of the newly added task classifier is set to $10$ times of the pre-trained parts. 
We set most of the hyper-parameters in SSL methods following recommendations of the original papers and empirically tune some hyper-parameters if necessary (e.g., when models do not converge).  Therefore, our results of SSL methods on UDA tasks are conservative estimates; however, such conservative estimates already justify the efficiency of SSL methods on UDA problems. 

Results on datasets of Office31, OfficeHome, VisDA-2017, and DomainNet are illustrated in Tables \ref{Tab:office31}, \ref{Tab:officehome}, \ref{Tab:visda}, and \ref{Tab:domainet}, respectively.   All SSL methods consistently improve over the Source Only baseline except the MixMatch on the most challenging pnt$\to$qdr task, justifying the effectiveness of SSL methods on UDA benchmarks.  Especially, the state-of-the-art SSL method of FixMatch outperforms existing UDA methods by more than $2.0$\% on the largest  UDA benchmark of DomainNet. 
Although MCC \cite{jin2020minimum} and Self-ensembling \cite{french2018selfensembling} are proposed as UDA methods for UDA tasks, they are variants of SSL methods. Specifically, Self-ensembling is a variant of the mean-teacher \cite{tarvainen2017mean} and MCC shares similar objectives to entropy minimization \cite{sslem} and self-training \cite{lee2013pseudo} (cf. Section \ref{Sec:similarity_methods}). 
Their superior results also imply the efficacy of SSL methods on UDA tasks.

\begin{table}
	\centering
	\caption{Inductive results on the large-scale DomainNet dataset (ResNet50). Transductive results are given in the appendices. } \label{Tab:domainet}
	\begin{small}
		\begin{tabular}{l|l|cccccc|c}
			\hline
			&Methods & clp$\to$inf & inf$\to$pnt & pnt$\to$qdr & qdr$\to$real & real$\to$skt & skt$\to$clp  & Avg. \\
			\hline
			& Source Only & 17.5$\pm$0.2 & 32.2$\pm$0.4 & 3.7$\pm$0.3 & 6.0$\pm$0.1 & 34.5$\pm$0.3 & 46.6$\pm$0.2 & 23.4 \\
			\hline
			\multirow{6}{*}{\centering\rotatebox{90}{UDA}} 
			& DANN \cite{dann} & 20.4$\pm$0.1 & 27.1$\pm$0.1 & \textbf{6.2}$\pm$0.3 & 15.6$\pm$0.4 & 40.0$\pm$0.3 & 48.0$\pm$0.3 & 26.2 \\
			& CDAN \cite{cada} & 20.4$\pm$0.2 & 28.9$\pm$0.1 & 3.4$\pm$0.1 & 16.0$\pm$0.2 & 42.2$\pm$0.1 & 49.6$\pm$0.5 & 26.7 \\
			& MCD \cite{mcd} & 19.8$\pm$0.2 & 34.1$\pm$0.1 & 3.4$\pm$0.2 & 16.6$\pm$0.1  & 41.3$\pm$0.1 & 50.4$\pm$0.1 & 27.6 \\
			& AFN \cite{xu2019larger} & 18.8$\pm$0.4 & 35.5$\pm$0.2 & \textbf{6.2}$\pm$0.4 & 16.5$\pm$0.1 & 37.9$\pm$0.1 & 51.2$\pm$0.3 & 27.7 \\
			& MDD \cite{zhang2019bridging} & 21.7$\pm$0.1 & 34.9$\pm$0.2 & 3.6$\pm$0.1 & \textbf{19.8}$\pm$0.2 & 44.6$\pm$0.1 & 54.0$\pm$0.1 & 29.7 \\
			\cline{2-9}
			& Self-ensembling \cite{french2018selfensembling} & 18.1$\pm$0.1 & 32.9$\pm$0.1 & 5.1$\pm$0.1 & 15.7$\pm$0.6 & 41.7$\pm$0.1 & 50.0$\pm$0.3  & 27.3 \\
			& MCC \cite{jin2020minimum} & 20.0$\pm$0.1 & 37.6$\pm$0.3 & 5.0$\pm$0.1 & 14.7$\pm$0.2 & 37.7$\pm$0.2 & 54.7$\pm$0.2 & 28.2 \\
			\hline
			\multirow{8}{*}{\centering\rotatebox{90}{SSL}} 
			& $\pi$-Model \cite{temporal_ensembling} & 18.0$\pm$0.1 & 34.2$\pm$0.1  & 5.3$\pm$0.1 & 11.5$\pm$0.1 & 35.0$\pm$0.1 & 48.6$\pm$0.2  &  25.4 \\
			& VAT \cite{vat} &  17.6$\pm$0.2 & 33.8$\pm$0.1 & 4.2$\pm$0.1 & 14.5$\pm$0.2 & 35.3$\pm$0.1 & 48.6$\pm$0.1 &  25.7 \\
			& Mean Teacher \cite{tarvainen2017mean} & 17.7$\pm$0.2 & 33.9$\pm$0.2 & 5.7$\pm$0.2 & 13.6$\pm$0.1 & 35.2$\pm$0.1 & 48.3$\pm$0.1 &  25.7 \\
			& Entropy mini. \cite{sslem} & 17.1$\pm$0.1 & 36.1$\pm$0.1 & 4.4$\pm$0.2 & 14.4$\pm$0.5 & 37.7$\pm$0.1 & 52.7$\pm$0.1 &  27.1  \\
			& MixMatch \cite{mixmatch} & 18.5$\pm$0.1 & 37.4$\pm$0.1 &  1.2$\pm$0.2 & 16.7$\pm$0.1 & 37.4$\pm$0.1 & 54.5$\pm$0.1 & 27.6   \\
			& Self-training \cite{lee2013pseudo} & 18.3$\pm$0.1 & 37.2$\pm$0.1 & 4.0$\pm$0.2 & 17.1$\pm$0.2 & 40.6$\pm$0.1 &  53.5$\pm$0.1 & 28.5 \\
			& Xie \emph{et al.} \cite{xie2019unsupervised} & 20.9$\pm$0.2 & 37.3$\pm$0.5 & 5.8$\pm$0.1 & 17.5$\pm$0.4 & 43.5$\pm$0.1 &  56.4$\pm$0.2 & 30.2 \\
			& FixMatch \cite{sohn2020fixmatch} & 21.8$\pm$0.2 & 40.7$\pm$0.5 & 4.4$\pm$0.3 & 18.4$\pm$0.4 & 45.9$\pm$0.2 & 59.2$\pm$0.2 & 31.7 \\
			\hline
			& MCC + Consistency & 19.5$\pm$0.1 & 39.0$\pm$0.1 & 4.2$\pm$0.2 & 16.2$\pm$0.1 & 38.4$\pm$0.1 & 55.3$\pm$0.1 & 28.8    \\
			& MDD + Consistency & \textbf{23.1}$\pm$0.6 & \textbf{41.6}$\pm$0.1 & 2.1$\pm$0.5 & 15.1$\pm$0.1 & \textbf{50.7}$\pm$0.2 & \textbf{61.9}$\pm$0.2 & \textbf{32.4} \\
			\hline
			& Oracle &  40.5$\pm$0.1 &  69.5$\pm$0.1 & 71.1$\pm$0.4 & 82.5$\pm$0.1 & 67.6$\pm$0.1& 75.4$\pm$0.2 & 67.8  \\
			\hline
		\end{tabular}
	\end{small}
\end{table}

We note that some works \cite{oliver2018realistic, yu2020multi} presented that SSL methods performed poorly when there was a \textit{class} distribution shift across labeled and unlabeled data; in contrast, we find that SSL methods perform well when there is a \textit{marginal} distribution shift (i.e., under the covariate shift assumption \ref{Equ:covariate_shift}), facilitating a deeper understanding of their application fields.



\subsection{Ablation study and analyses} \label{Sec:analyses}

\textbf{Combining UDA and SSL methods for UDA tasks.}
We find that state-of-the-art UDA methods could be boosted with SSL techniques. Specifically, we combine the popular consistency regularization in SSL \cite{xie2019unsupervised,sohn2020fixmatch} with UDA methods of MCC \cite{jin2020minimum} and MDD \cite{zhang2019bridging}, leading to boosted results and the new state of the art, as illustrated in Tables \ref{Tab:office31}, \ref{Tab:officehome}, \ref{Tab:visda}, and \ref{Tab:domainet}. The boosted results also justify the efficacy of SSL techniques on UDA tasks. The combining strategies are detailed in the appendices.


\begin{figure}
	\centering
	\subfigure[$250$ labeled samples] {
		\label{Fig:uda_on_ssl}
		\includegraphics[width=0.23\textwidth]{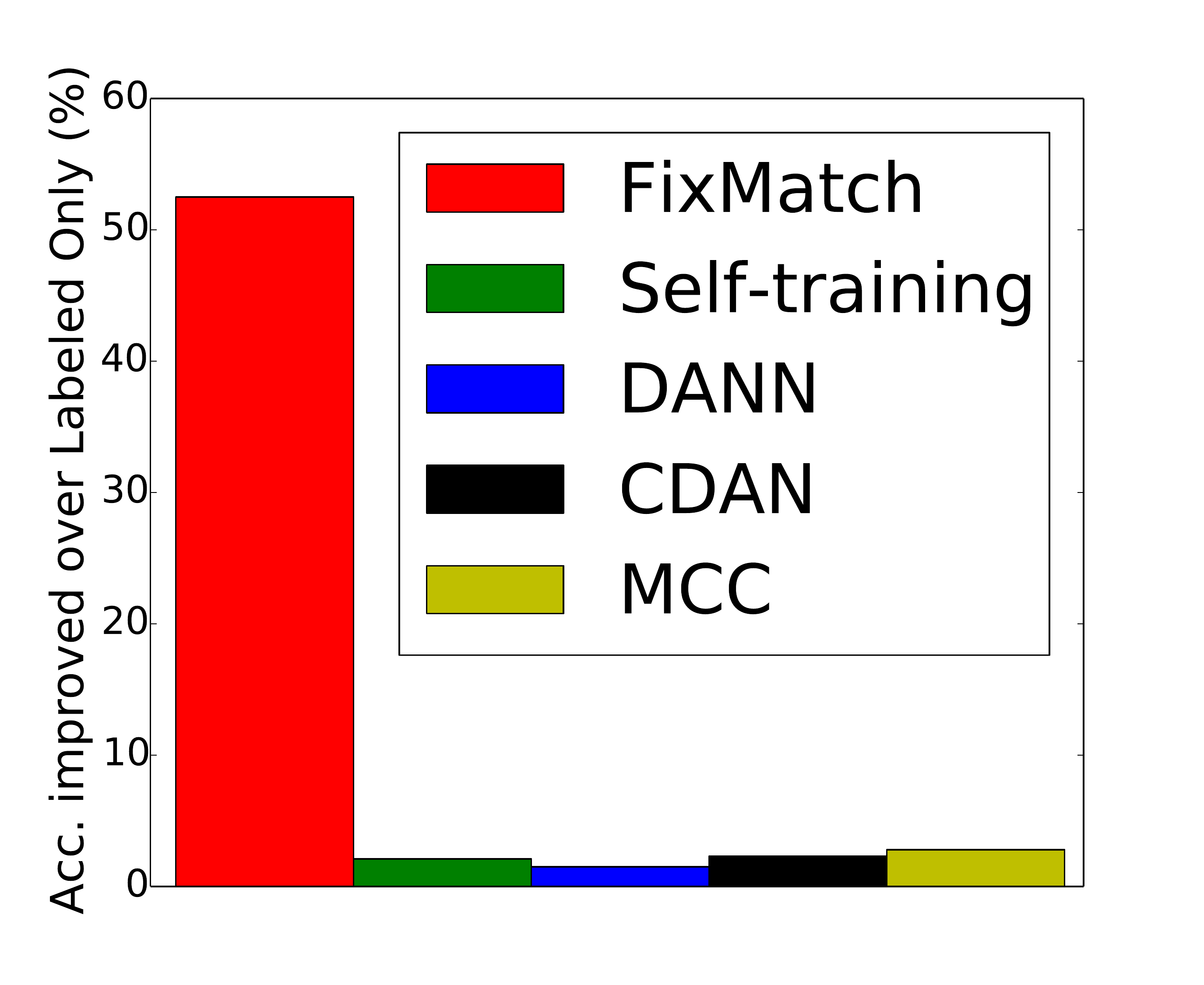}
	}
	\subfigure[$4000$ labeled samples] {
		\label{Fig:uda_on_ssl_4000}
		\includegraphics[width=0.23\textwidth]{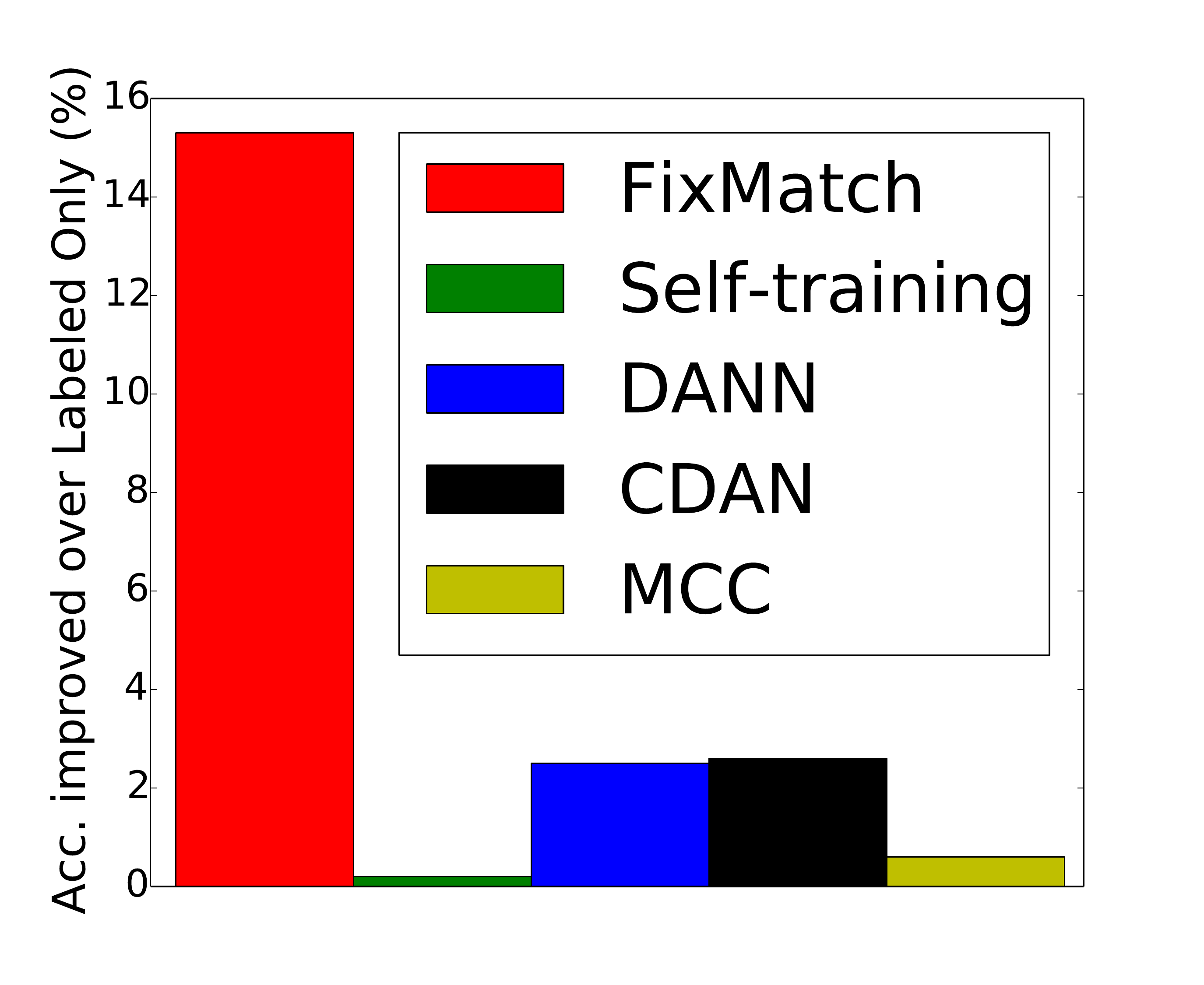}
	}
	\subfigure[SSL implementations] {
	\label{Fig:ssl_on_G_only}
	\includegraphics[width=0.23\textwidth]{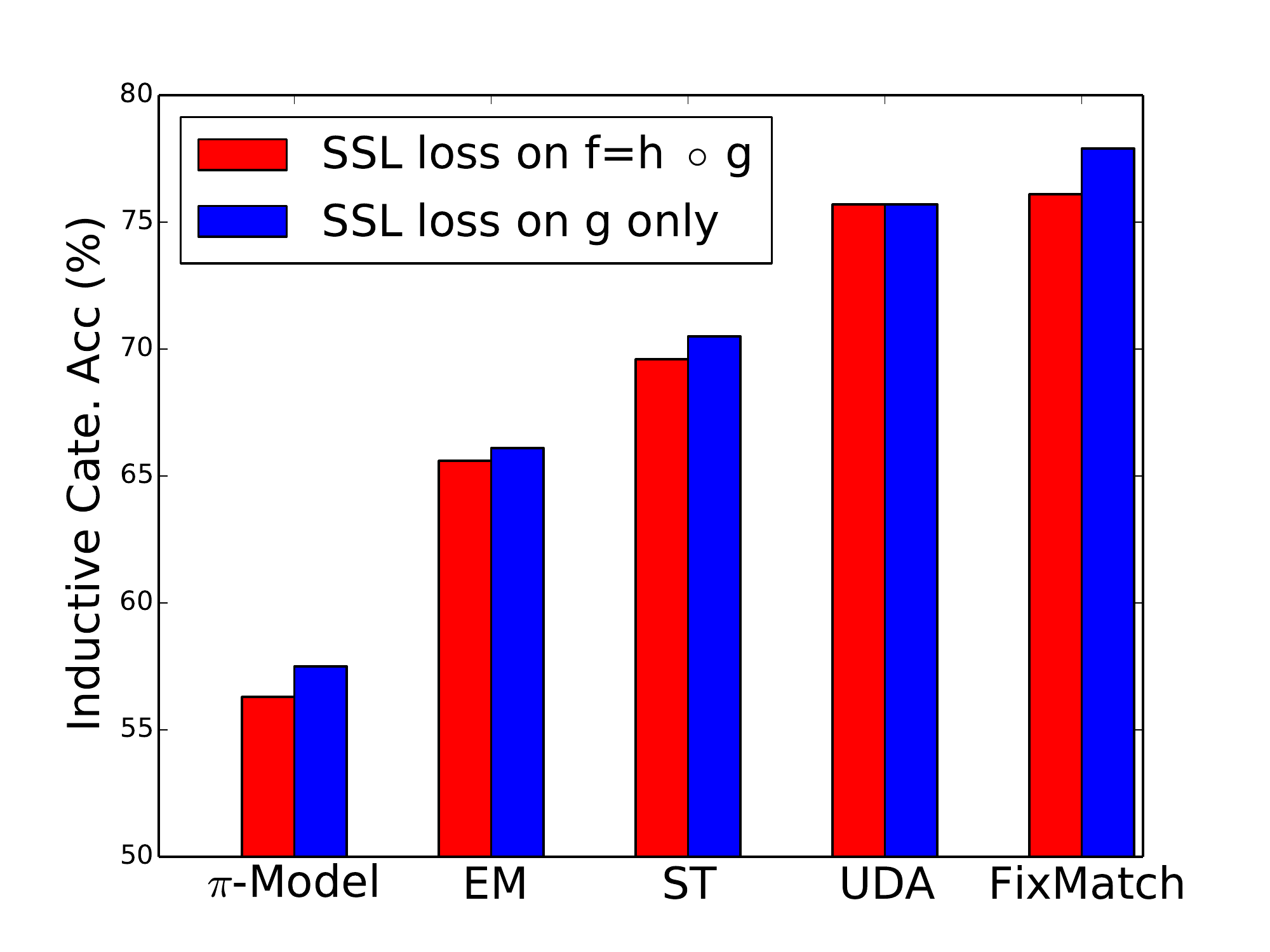}
	}
	\subfigure[Convergence] {
		\label{Fig:convergence_visda}
		\centering
		\includegraphics[width=0.23\textwidth]{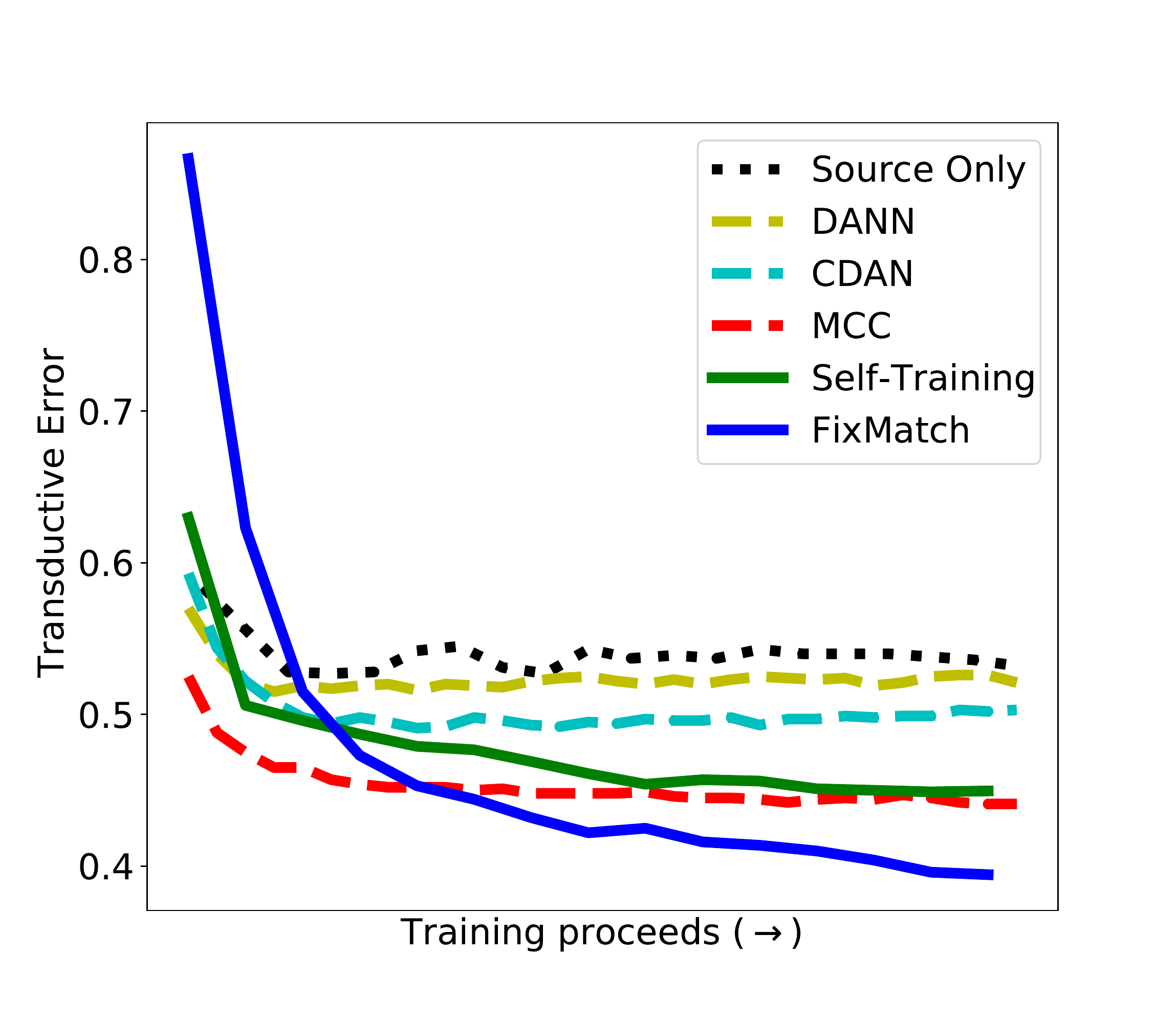}
	} 
	\caption{ 
		 (a)-(b): Performance improvement over the Labeled Only on the Cifar10 dataset with (a) $250$ and (b) $4000$ labeled samples in the SSL task. (c): Inductive results on VisDA-2017 dataset with settings of `SSL loss on $f=h \circ g$' (\ref{Equ:general_objective}) and `SSL loss on $g$ only' (cf. appendices for details). 
		 We abbreviate entropy minimization and self-training to EM and ST, respectively. (d): Convergence performance on the skt$\to$clp task of the DomainNet dataset. }
\end{figure}

\begin{figure}
	\centering
	\subfigure[VisDA-2017] {
		\label{Fig:a_dis}
		\centering
		\includegraphics[width=0.225\textwidth]{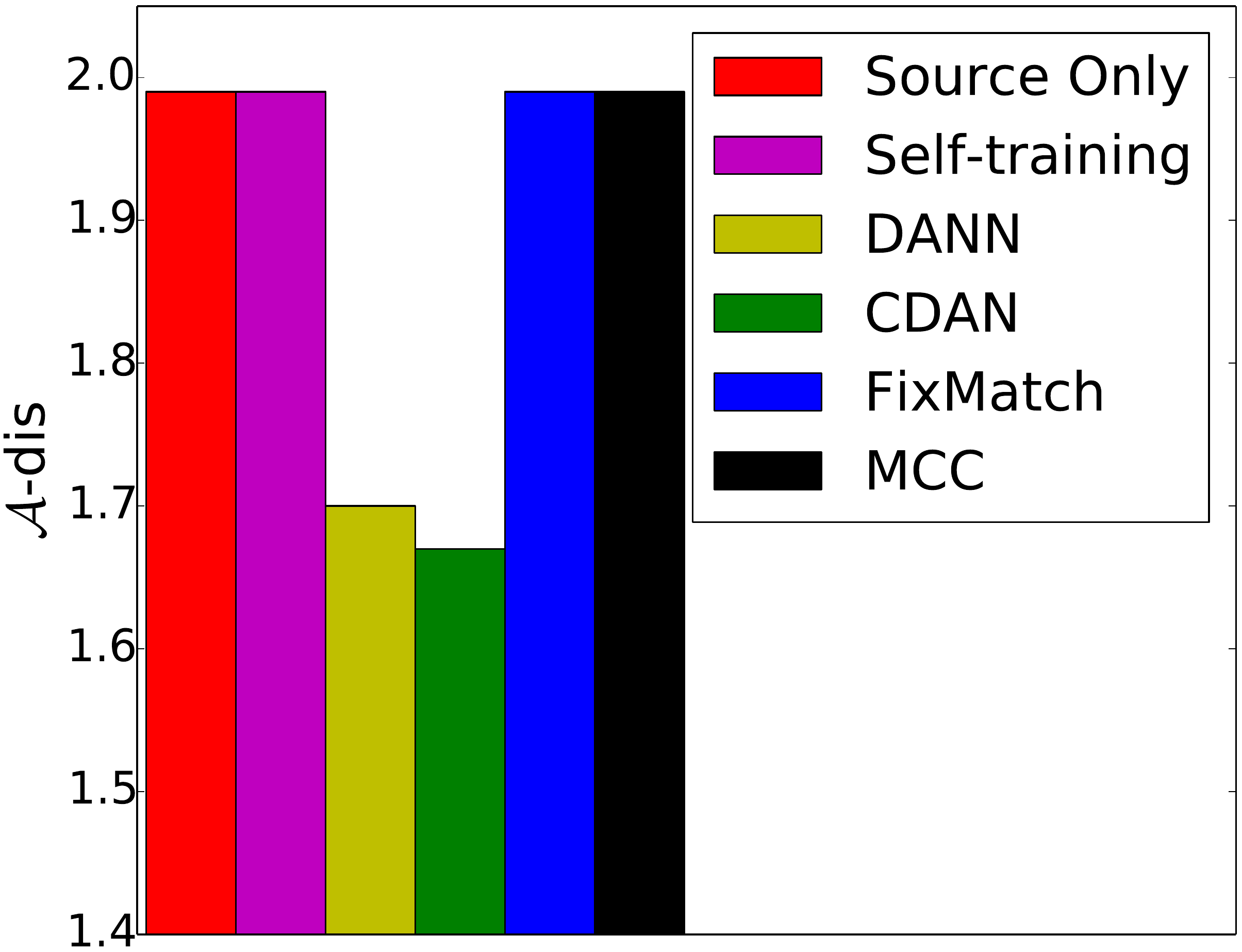}
	} 
	\subfigure[A$\to$W] {
		\label{Fig:a_dis_office31}
		\centering
		\includegraphics[width=0.225\textwidth]{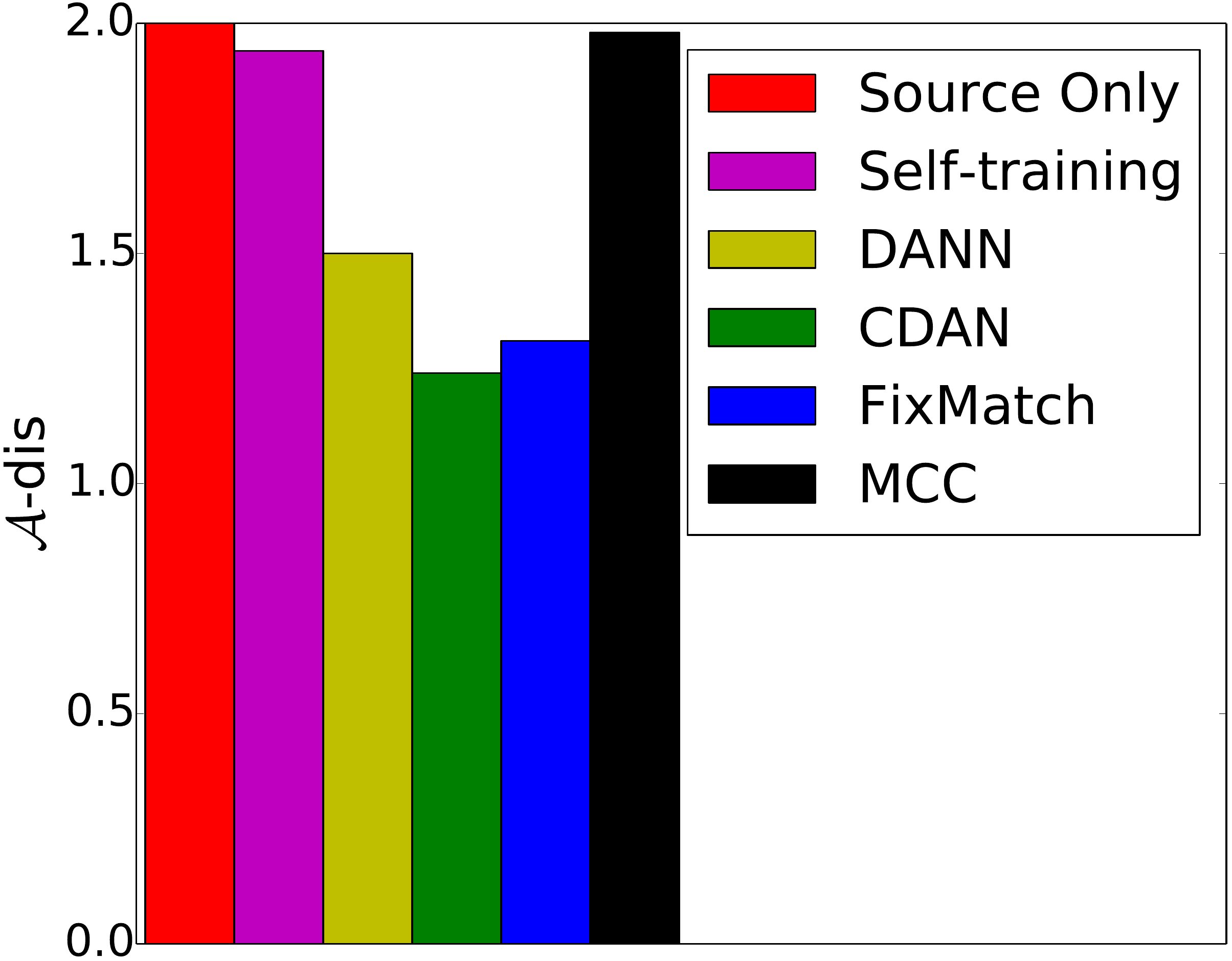}
	} 
	\subfigure[OfficeHome] {
		\label{Fig:data_processing_home}
		\centering
		\includegraphics[width=0.225\textwidth]{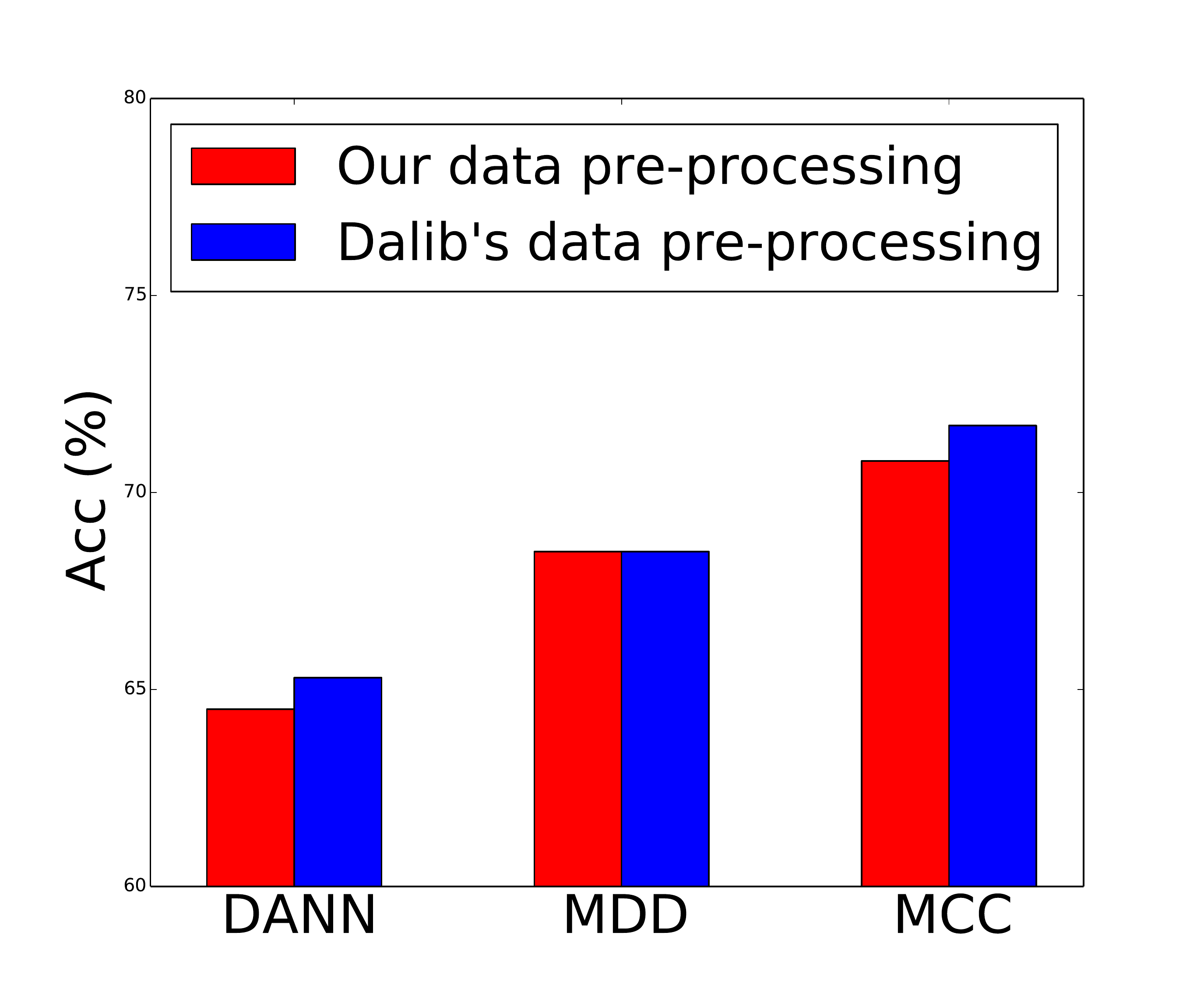}
	} 
	\subfigure[DomainNet] {
		\label{Fig:data_processing_domainnet}
		\centering
		\includegraphics[width=0.225\textwidth]{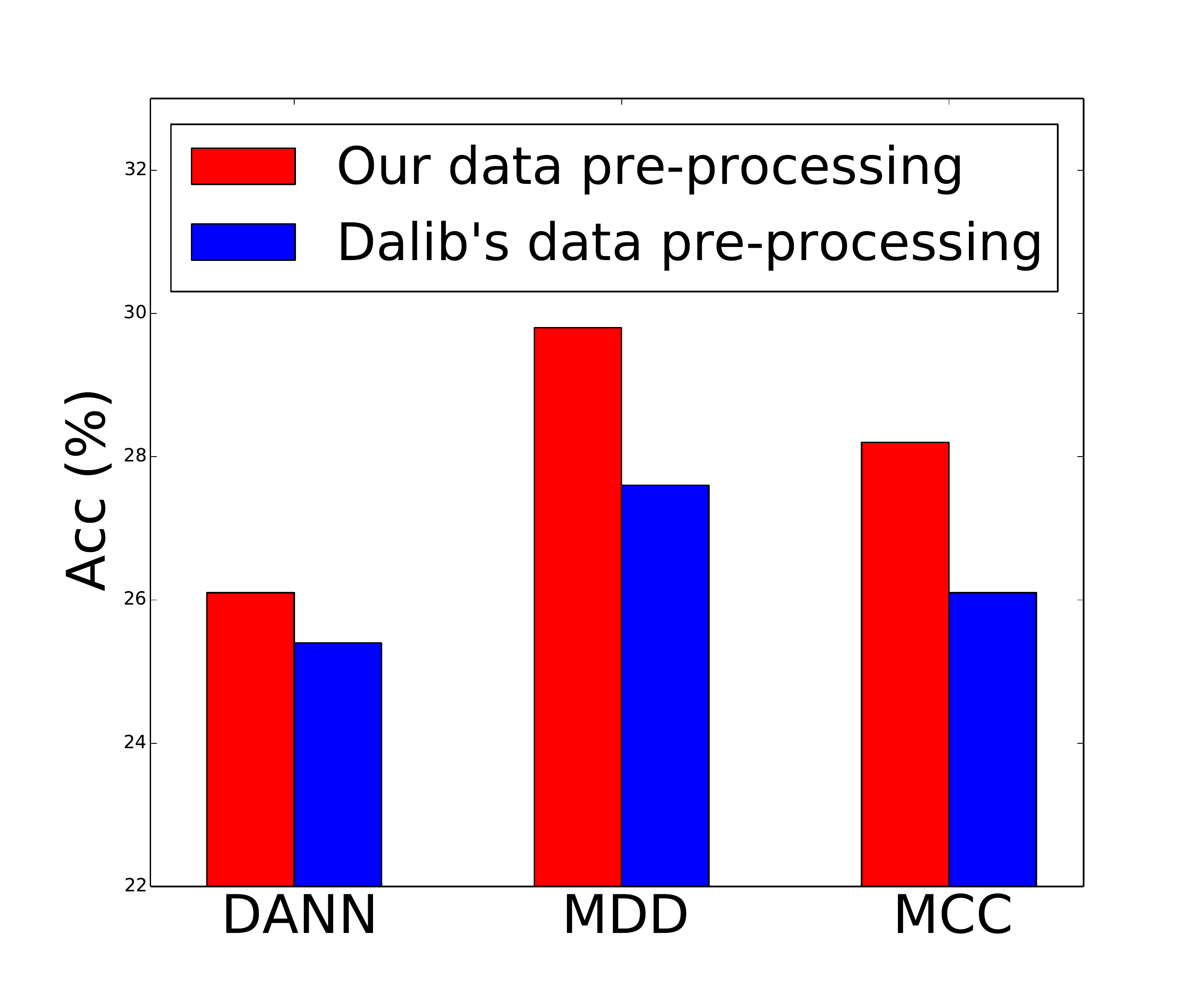}
	} 
	\caption{(a)-(b): $\mathcal{A}$-distance between source and target features on the  (a) VisDA-2017 dataset and (b) A$\to$W task of Office31 dataset, where values are rounded to the level of $0.01$.  (c)-(d): Results with different data pre-processing strategies on the datasets of (c) OfficeHome and (d) DomainNet. }
\end{figure}

\textbf{Employing UDA methods for SSL tasks.}
Given the close relationship between UDA and SSL and the superior results of SSL methods on UDA tasks, one may consider employing UDA methods on SSL tasks. 
As illustrated in Figures \ref{Fig:uda_on_ssl} and \ref{Fig:uda_on_ssl_4000}, UDA methods of DANN \cite{dann}, CDAN \cite{cada} and MCC \cite{jin2020minimum} marginally outperform the `Labeled only' baseline, presenting  limited efficacy on SSL tasks. We also attempt to combine the state-of-the-art SSL method of FixMatch \cite{sohn2020fixmatch} with classical UDA techniques (e.g., adversarial distribution minimization in \cite{dann,cada}) for SSL tasks, but we observe no improvement over vanilla FixMatch.

We note that a technique termed distribution alignment is adopted in the recent SSL method \cite{berthelot2019remixmatch}; however, it is different from the popular distribution alignment in UDA.  Berthelot \emph{et al.} \cite{berthelot2019remixmatch} aim to align the \textit{marginal class distribution} across labeled and unlabeled data, which is motivated by the input-output mutual information maximization objective. On the contrary, the popular distribution alignment in UDA aligns the \textit{marginal feature distribution} across labeled and unlabeled data. 

\textbf{Applying SSL losses on the feature extractor $g$ only for UDA tasks.}
Considering the domain shift, researchers typically apply the entropy minimization loss on the feature extractor $g$ only for UDA tasks \cite{zhang2018importance,symnets}. We analyze whether this benefit could generalize to other SSL regularizations (\ref{Equ:regularization}).  
 As illustrated in Figure \ref{Fig:ssl_on_G_only}, better results of SSL methods on UDA benchmarks are typically achieved by applying  SSL losses on the feature extractor $g$ only (cf. the appendices for details), which is adopted as the default implementation of SSL methods for UDA tasks in this paper.

%




\textbf{Convergence performance.}  We illustrate the convergence performance on the skt$\to$clp task of the DomainNet dataset in Figure \ref{Fig:convergence_visda}.
UDA and SSL methods both converge smoothly.

\textbf{Domain divergence.}
DANN \cite{dann} and CDAN \cite{cada}, which are designed for the divergence minimization, consistently and significantly minimize the distribution divergence on various tasks, as illustrated in Figures \ref{Fig:a_dis} and \ref{Fig:a_dis_office31}. 
On the contrary, most SSL methods and their variants marginally reduce the domain divergence; although FixMatch considerably reduces the domain divergence on the A$\to$W task, it hardly reduces the divergence on the VisDA-2017 dataset, a task with larger domain divergence, which distinguishes it from the seminal UDA methods (e.g., DANN and CDAN). In practice, SSL methods and their variants currently achieve the state of the art, as shown in Tables \ref{Tab:office31}, \ref{Tab:officehome}, \ref{Tab:visda}, and \ref{Tab:domainet}.  Similar to \cite{chen2019transferability,zhao2019learning}, this empirical evidence challenges the dominant position of divergence minimization in UDA studies and suggests another promising UDA solution of SSL methods. 

\textbf{Data pre-processing.}
 As illustrated in Figures \ref{Fig:data_processing_home} and \ref{Fig:data_processing_domainnet},
 results on UDA tasks are significantly influenced by the data pre-processing strategies; for example, different data pre-processing strategies lead to more than $2$\% accuracy divergence for the MCC method on the DomainNet dataset, which is even larger than its improvement over most compared methods.  Thus, we promote that researchers should
 compare different methods on UDA tasks with the same data pre-processing, which is less concerned in previous UDA studies. We detail the data pre-processing strategies in the appendices.

\section{Conclusions and discussions} \label{Sec:conclusion}
	In this work, we bridged UDA and SSL by revealing that SSL is a special case of UDA problems. By adapting eight representative SSL methods on UDA benchmarks, we showed that SSL methods were strong UDA learners. Specifically, the state-of-the-art SSL method of FixMatch outperformed existing UDA methods by more than $2.0$\% on the challenging DomainNet benchmark, and the new state of the art could be achieved by empowering UDA methods with SSL techniques. We thus promoted that SSL methods should be employed as baselines in future UDA studies. Our work bridged existing tasks of UDA and SSL; thus, it inherited their negative impacts (e.g., abused models).
	
	However, in experiments, we conducted the model selection (e.g., choosing training checkpoints) according to results on labeled target data, which is actually a common problem in the UDA community; although the upper bound performance of SSL and UDA methods \cite{dalib} were fairly compared, we expect to compare them on UDA tasks with appropriate model selection strategies in the future.

{\small
	\bibliographystyle{plain}
	\bibliography{egbib}
}

\end{document}